\documentclass[twocolumn]{article}
\usepackage{spconf,amsmath,graphicx}
\usepackage{dblfloatfix}
\usepackage[normalem]{ulem}

\usepackage{CJKutf8}
\usepackage{url}
\usepackage{subcaption}
\usepackage{hhline}
\usepackage{algorithm}
\usepackage{algorithmic}
\usepackage{amssymb}
\usepackage{amsmath}    
\usepackage{multirow}
\usepackage{color}
\usepackage{booktabs}
\usepackage{balance}

\newcommand{\zhenk}[1]{{\color{black}{#1}}}

\newcommand{\bftab}{\fontseries{b}\selectfont}

\usepackage{bm}

\newcommand{\bh}{\bm{h}}

\newcommand{\bs}{\bm{s}}
\newcommand{\bt}{\bm{t}}

\newcommand{\bx}{\bm{x}}
\newcommand{\by}{\bm{y}}
\newcommand{\bz}{\bm{z}}

\newcommand{\bC}{\bm{C}}

\newcommand{\bU}{\bm{U}}

\newcommand{\bW}{\bm{W}}

\newcommand{\calF}{{\mathcal{F}}}
\newcommand{\calG}{{\mathcal{G}}}

\newcommand{\be}{\begin{align}}
\newcommand{\ee}{\end{align}}
\newcommand{\bee}{\begin{align*}}
\newcommand{\eee}{\end{align*}}

\newcommand{\matrixb}{\left[ \begin{array}}
\newcommand{\matrixe}{\end{array} \right]}

\newcommand{\argmin}{\operatornamewithlimits{\arg \min}}

\usepackage[font=small,labelfont=bf]{caption}

\title{Sparsification via compressed sensing \\ for Automatic Speech Recognition}
\name{Kai Zhen$^{1}$\sthanks{This work was conducted during Kai’s internship in Amazon Pittsburgh, PA, USA.}, Hieu Duy Nguyen$^{2}$, Feng-Ju Chang$^{2}$, Athanasios Mouchtaris$^{2}$, and Ariya Rastrow$^{2}$ }

\address{$^{1}$Indiana University Bloomington  \\	$^{2}$Alexa Machine Learning, Amazon, USA\\
	\texttt{zhenk@indiana.edu}, ~\texttt{\{hieng, fengjc, mouchta, arastrow\}@amazon.com}}

\begin{document}
	
	\topmargin=0mm
	\maketitle
	\ninept

	\begin{abstract}
		In order to achieve high accuracy for machine learning (ML) applications, it is essential to employ models with a large number of parameters. Certain applications, such as Automatic Speech Recognition (ASR), however, require real-time interactions with users, hence compelling the model to have as low latency as possible. Deploying large scale ML applications thus necessitates model quantization and compression, especially when running ML models on resource constrained devices. For example, by forcing some of the model weight values into zero, it is possible to apply zero-weight compression, which reduces both the model size and model reading time from the memory. In the literature, such methods are referred to as sparse pruning. The fundamental questions are when and which weights should be forced to zero, i.e. be pruned. 
		In this work, we propose a compressed sensing based pruning (CSP) approach to effectively address those questions. By reformulating sparse pruning as a sparsity inducing and compression-error reduction dual problem, we introduce the classic compressed sensing process into the ML model training process. Using ASR task as an example, we show that CSP consistently outperforms existing approaches in the literature.
	\end{abstract}

	\section{Introduction}
	\label{sec:intro}
		
	Automatic Speech Recognition (ASR) is an important component of a virtual assistant system. The main focus of ASR is to convert users' voice command into transcription, based on which further processing will act upon. Recently, end-to-end (E2E) approaches have attracted much attention due to their ability of directly transducing audio frame features into sequence outputs \cite{prabhavalkar2017comparison}. Without explicitly imposing/injecting domain knowledge and manually tweaking intermediate components (such as lexicon model), building and maintaining E2E ASR system is much more efficient than a hybrid deep neural network (DNN)-Hidden Markov Model (HMM) model.
	
	In order to provide the best user experience, an ASR system is required to achieve not only high accuracy but also small user-perceived latency. This motivates the trend of moving the processing from Cloud/remote servers to users' device to reduce the latency further. More often than not, the hardware limitations impose strict constraints on the model complexity \cite{han2015deep,punjabi2020streaming_amazon}. Firstly, the hardware may only support integer arithmetic operations for run-time inference, which necessitates model quantization. Secondly, a hardware often performs multiple tasks supported by different models in sequence. The hardware, with limited memory size, thus needs to move multiple models in and out of the processing units. 
	%Therefore, both memory storage and bandwidth of the hardware limit the model complexity.
	%The model complexity is therefore restricted by both memory storage and model bandwidth of the hardware.
	
	To compress the model for the hardware, two widely applied methods are (a) model selection/structured pruning, i.e. choosing a model structure with pruned layers/channels and small performance degradation \cite{cao2019efficient,wang2019acceleration}, and (b) zero-weight compression/sparse pruning, i.e. pruning small-value weights to zero \cite{zhu2017prune,narang2017exploring}. Model selection differs from sparse pruning in that it deletes entire channels or layers, showing a more efficient speedup during inference, yet with a  more severe performance degradation \cite{cao2019efficient,wang2019acceleration}. These two types of methods are usually complementary: after being structurally pruned, a model can also undergo further zero-weight compression to improve the inference speed.
	In this study, we focus on sparse pruning, targeting at lowering the memory storage and bandwidth requirement as it largely contributes to the latency for on-device ASR models.
	
	A na\"{\i}ve approach for sparse pruning is to push the weight values smaller than a threshold  to zero after training, which  often leads to significant performance degradation \cite{zhu2017prune}. To mitigate this problem, Tibshirani \textit{et. al.} applied \textit{LASSO} regularization to penalize large-value model weights \cite{tibshirani2005sparsity}. The drawbacks are twofold: firstly, it does not exert an explicit specification of the target sparsity level; secondly, it is subject to the gradient vanishing issue when the model has more and more layers. Gradual pruning approach \cite{zhu2017prune} resolves those concerns by defining a sparsity function that maps each training step to a corresponding intermediate sparsity level. During the model training, the pruning threshold is adjusted gradually according to the function to eventually reach the target sparsity level. However, gradual pruning assumes that pruning the values smaller than the threshold will lead to the least degradation, which is heuristic and sub-optimal. Consequently, gradual pruning techniques provide a guidance on ``when to prune and by how much'', but a less satisfying answer for ``which (weights) to prune'', thus leading to inefficiency.
	
	In this work, we propose a compressed sensing (CS)-based pruning method, referred to as CSP subsequently, that is sparse-aware and addresses both ``\textit{when to prune}'' and ``\textit{which to prune}''. CSP reformulates the feedforward operations in machine learning architectures, such as Long Short Term Memory (LSTM) or Fully-Connected (FC) cells, as a sensing procedure with the inputs and hidden states being random sensing matrices. Under that perspective, a sparsification process is to enhance the sparsity and reduce the compression error, due to pruning, simultaneously. Following \cite{donoho2006compressed}, we adopt the $\ell_1$ regularization to enforce sparsity and the $\ell_2$ regularization to mitigate the compression loss, and reformulate the sparsification procedure as an optimization problem. We demonstrate the effectiveness of our method by compressing recurrent neural network transducer (RNN-T), one of the E2E ASR models. The RNN-T model is sparsified via a hybrid training mechanism in which CSP is conducted during the feedforward stage, along with the back propagation for the global optimization. Our proposed method constantly outperforms the state-of-the-art gradual pruning approach in terms of the word error rate (WER) under all settings. In particular, with a sparisty ratio of $50\%$  where half of the weights are $0$,  CSP yields little to no performance degradation on LibriSpeech dataset. 
	
	The rest of the paper is structured as follows. In Sec. \ref{sec:relatedwork}, we briefly review related pruning methods. Our CSP method is introduced in Sec. \ref{sec:approach}. Sec. \ref{sec:exp} describes our experiment setup and results. Finally, we conclude our work with some remarks in Sec. \ref{sec:conclusion}.
	
\section{Related work}
	\label{sec:relatedwork}

	One of the most straightforward approaches for sparse-aware training is applying $\ell_k$ regularization, where $k=0, 1, 2,$ etc. There has been a rich literature in comparing various forms of sparsity regularizers. Consider the model training: $\mathbb{W}\leftarrow\argmin_{\mathbb{W}} \mathcal{L}_{\text{accuracy}}(\mathbb{W}) + ||\mathbb{W}||_1 $, where $\ell_1$ norm is used on the regularization. The fundamental idea is to optimize the model prediction while penalizing large weight values. In DNN model compression, the regularization is usually implemented as an extra loss term for the training.
	
	This training-aware sparsity regularization leads to promising pruning results especially for convolutional neural networks (CNN) \cite{lou2018sparse} with residual learning techniques \cite{he2016deep,huang2017densely}, but may not apply well to models employing recurrent neural network (RNN) components such as LSTM. The error due to a global sparsity constraint $\ell_1$ will be propagated to all time steps. Additionally, such drawback is much more severe for architectures, such as RNN-T, which contains feedback loop from one part of the model to the others.
	
	Another well-known, state-of-the-art, pruning method for ML models is gradual-pruning \cite{zhu2017prune}. This method does not resort to $\ell_1$ regularization for sparsity, but dynamically updates the pruning threshold during model training, as is indicated by its name.
	To answer the question "when to prune", the authors defines a sparsity function parametrized by the  target sparsity $\bs_f$ at $\bt_n$ step with an initial pruning step $\bt_0$. Concretely, at training step $\bt$, the pruning threshold is adjusted to match the sparsity $\bs_t$ calculated in Eq. \ref{eq:sp}. The main complication is to  adjust the pruning procedure such that the model weights are relatively converged and the learning rate is still sufficiently large to reduce the pruning-induced loss.
	
	\begin{equation}
	\label{eq:sp}
	\bs_t=\bs_f * \left( 1 - \left( 1-\frac{\bt-\bt_0}{\bt_n-\bt_0} \right)^3 \right)
	\end{equation}
	
	One concern is that finding an optimal setup for these hyperparameters can be hard without going through a rigorous ablation analysis. Furthermore, with gradual pruning, gradient-updating backpropagation is the only mechanism to limit the degradation.
	Most importantly, gradual pruning only addresses the question ``\textit{when to prune}'' but not ``\textit{which (weights) to prune}''. At each time step, the weights below the (gradually increased) threshold are to be pruned. This is based on the premise that the smaller the weight, the less important it is, which is heuristic and sub-optimal.

\section{Compressed Sensing based pruning}
	
	\label{sec:approach}
	\subsection{Adapting Compressed Sensing for Sparse Pruning}
	CS aims to compressing potentially redundant information into a sparse representation and reconstructing it efficiently  \cite{yuan2020image, qiao2020deep}, which has facilitated a wide scope of engineering scenarios, such as medical resonance imaging (MRI) and radar imaging.
	For example, in MRI \cite{lustig2008compressed}, high resolution scanned images are generated per millisecond or microsecond, leading to significant storage cost and transmission overhead. CS learns a sparse representation of each image with which, during the decoding time, CS can recover the reference image almost perfectly. Assume an image compression task with a reference image $\bx \in \mathbb{R}^{n}$,
	where $\bx$ is usually decomposed into an orthogonal transformation basis $\psi$ and the activation $\bs$, as $\bx = \psi*\bs$.
	Given that $\bs$ satisfies $\mathcal{K}$-sparse property, CS is capable of locating those $\mathcal{K}$ salient activation elements.
	Concretely, CS  introduces a sensing matrix $\phi$ to project $\bx$ into $\by$, as $\by= \phi*\bx$. In \cite{donoho2006compressed,candes2006robust}, it has been proved that by optimizing the  $\ell_2$ loss in the sensing dimensionality while exerting the $\ell_1$ norm regularizer to $\bs$, a $\mathcal{K}$-sparse solution of $\bs$, denoted as $\hat{\bs}$, can be found in polynomial time. Consequently, $\psi*\hat{\bs}$ can estimate the original image $\bx$ with high fidelity and relatively small latency.

	In this work, we investigate the effectiveness of CS based pruning for ML models. We consider the ASR task, i.e. converting audio speech to transcriptions, using an RNN-T architecture. Due to the space constraint, we only describe the transformation of LSTM cell, which is a major building block in various E2E ASR models. 
	It is straightforward to extend the transformations to other architectures/layers like the fully-connected (FC) network and CNN. 
	
	Consider a vanilla LSTM cell: the element-wise multiplication between the input at time step $t$, $\bx^{(t)}$, and kernels is given in Eq. \ref{eq:lstm_op1}, while that of hidden states from the previous step $\bh^{(t-1)}$ and recurrent kernels is in Eq. \ref{eq:lstm_op2}. Here, $\bW_f,\bW_c,\bW_i,$ and $\bW_o$ (correspondingly $\bU_f,\bU_c,\bU_i,$ and $\bU_o$) denote the kernels (correspondingly recurrent kernels) weights of the cell ($c$), the input gate ($i$), output gate ($o$), and forget gate ($f$), respectively. All gating mechanisms to update the cell and hidden states are encapsulated in Eq. \ref{eq:lstm_op3}, where $\bC^{(t)}$ is the cell state vector at time $t$ and $\calG$ denotes the transformation of $\bz_x, \bz_h, \bC^{(t-1)}$ into $\bC^{(t)}, \bh^{(t)}$. The bias terms are omitted for ease of presentation.
	\begin{align}
	\label{eq:lstm_op1}&\bz_x = [\bW_f,\bW_c,\bW_i,\bW_o]\odot[\bx^{(t)}, \bx^{(t)}, \bx^{(t)}, \bx^{(t)}]\\
	\label{eq:lstm_op2}&\bz_h = [\bU_f,\bU_c,\bU_i,\bU_o]\odot[\bh^{(t-1)}, \bh^{(t-1)}, \bh^{(t-1)}, \bh^{(t-1)}]\\
	\label{eq:lstm_op3}&[\bC^{(t)}, \bh^{(t)}] = \calG(\bz_x, \bz_h, \bC^{(t-1)}),
	\vspace{-6mm}
	\end{align}
	
	\begin{figure*}[t]
		%\label{fig:box}
		\centering
		\includegraphics[scale=0.4]{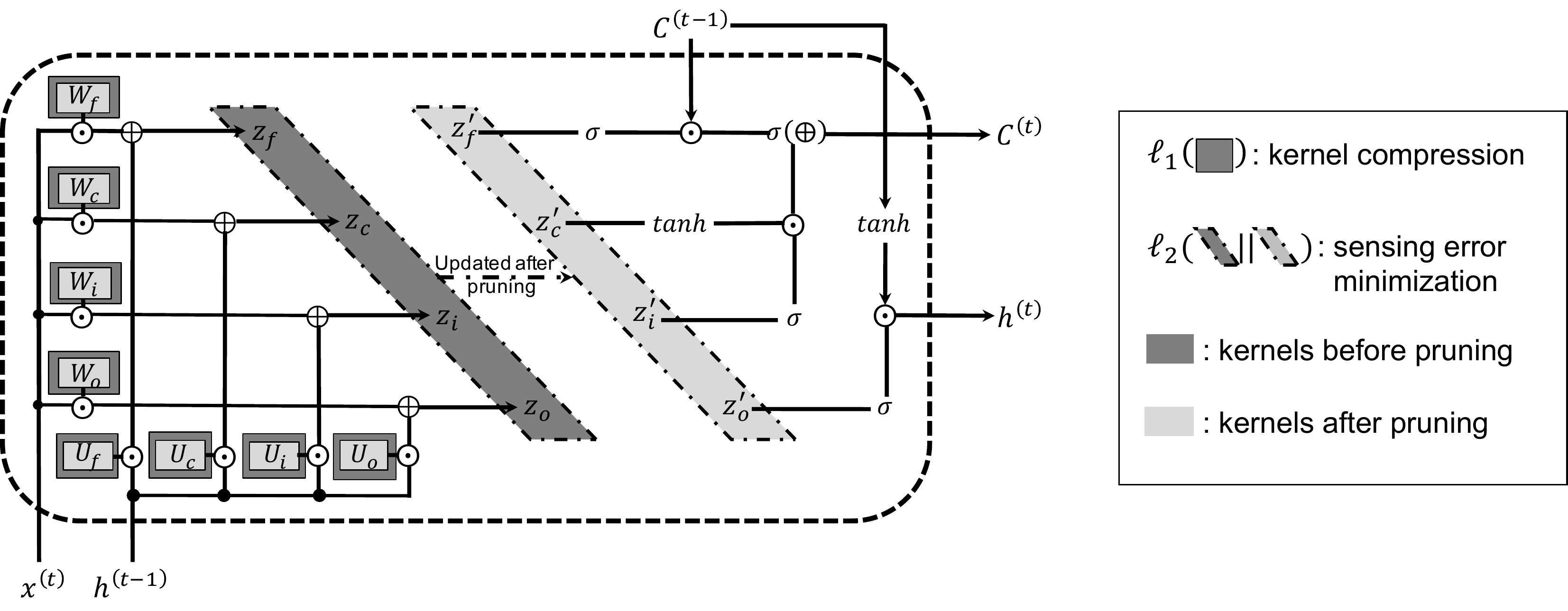}
		\caption{A CSP-LSTM cell with the local sparse optimizer}
		\label{fig:cell}
		\vspace{-4mm}
	\end{figure*}

	To prune all kernel  weights (denoted as $\mathbb{W}=[\bW_f, \bW_c, \bW_i, \bW_o]$) in LSTM cells,  we adapt and reformulate a CS-like pruning procedure by adopting $\ell_1$ regularization for sparsity-inducing and $\ell_2$ regularization for compression-error reduction, as outlined in Fig. \ref{fig:cell}. The procedure starts by a midway inference to collect the activation inputs $[\bz_x, \bz_h]$ as in Eq. \ref{eq:lstm_op1} and Eq. \ref{eq:lstm_op2}. When those kernels are sparsified, the activation inputs will be consequently updated as $[\bz_x', \bz_h']$ (see Fig. \ref{fig:cell}). The goal is to sparsify and prune the kernels while preserving the value of $[\bz_x, \bz_h]$ to minimize the pruning-induced loss. To that end, the $\ell_1$ regularizer is applied to the input kernels while the $\ell_2$ regularizer controls the reconstruction loss on $\bz_x$.  Hence, our CS solver is embedded in a local optimizer triggered periodically by feedforward steps in a stochastic manner. 
    \zhenk{As illustrated in the restricted isometry property (RIP), the sensing matrix $\phi$ is expected to be random for an accurate signal reconstruction \cite{candes2006robust}. The proposed CS solver satisfies the RIP with high probability, since the sensing matrices (the input $\bx^{(t)}$ and hidden state $\bh^{(t)}$) in LSTM cells vary at different time steps.}
    
    %For LSTM cells, the sensing matrices (the input $\bx^{(t)}$ and hidden state $\bh^{(t)}$) satisfy the RIP with high probability, as they vary at different time steps.
    
	%Note that $\bx^{(t)}$ varies at different time steps and batches, which can be considered as a sensing matrix (similar to the $\phi$ matrix in the discussion of CS above) that satisfies the RIP with high probability, as experimentally verified Sec. \ref{sec:exp}. 
	
	Our CS solver is kernel-wise, i.e. in each LSTM layer, there is one CS solver for each of the input-kernel and recurrent-kernel sparsification process. A general CS loss for the local optimizer is defined in Eq. \ref{eq:cs}.  By adjusing the sensing coefficient, the local optimizer is capable of calibrating the model to the target sparsity by balancing the $\ell_1$ and  $\ell_2$  regularizers, 
	\begin{equation}
	\label{eq:cs}
	\mathcal{L}_{CS}(\mathbb{W}, \by, \bh)=\lambda|\mathbb{W}|_1+||\by-\mathbb{W}\odot\bh||_2,
	%\end{aligned}
	\vspace{-1mm}
	\end{equation}
where the hyperparameter $\lambda$ in  Eq. \ref{eq:cs} is referred to as the sensing coefficient subsequently. $\mathbb{W}$, $\bh$, and $\by$ denote kernel weights, inputs, and activation input, respectively. 
	To remove the manual-tuning, we  dynamically update $\lambda$ via Eq. \ref{eq:lambda}. 

	\begin{equation}
	\label{eq:lambda}
	\lambda \leftarrow \left \{
	\begin{aligned}
	&\max(\lambda_{\text{lower}}, \lambda - \epsilon), && \text{if}\ \bs>\bs_t \\
	&\min(\lambda_{\text{upper}}, \lambda + \epsilon), && \text{otherwise}.
	\end{aligned} \right.
	\end{equation}

	During training, if the actual sparsity overshoots the target sparsity at a certain step, we reduce $\lambda$ by $\epsilon$. Otherwise, the $\lambda$ is increased by $\epsilon$ instead. In our setup, $\lambda$ is constrained between $\lambda_{\text{lower}}=0.001$ and $\lambda_{\text{upper}}=1.0$ with $\epsilon$ being 0.005. The pruning threshold, $\rho$, initialized as $0.002$, is also updated as in \cite{zhu2017prune} to adjust the sparsity.
	Algorithm \ref{algo:1} summarizes our proposed CSP procedure for input kernels. The CSP for recurrent kernels $\mathbb{U}=[\bU_f, \bU_c, \bU_i, \bU_o]$ is executed similarly and is omitted for brevity.
	
	\subsection{Hybrid Sparse-Aware Training Scheme}
\begin{figure}[t]
		%\label{fig:box}
		\centering
		\includegraphics[scale=0.32]{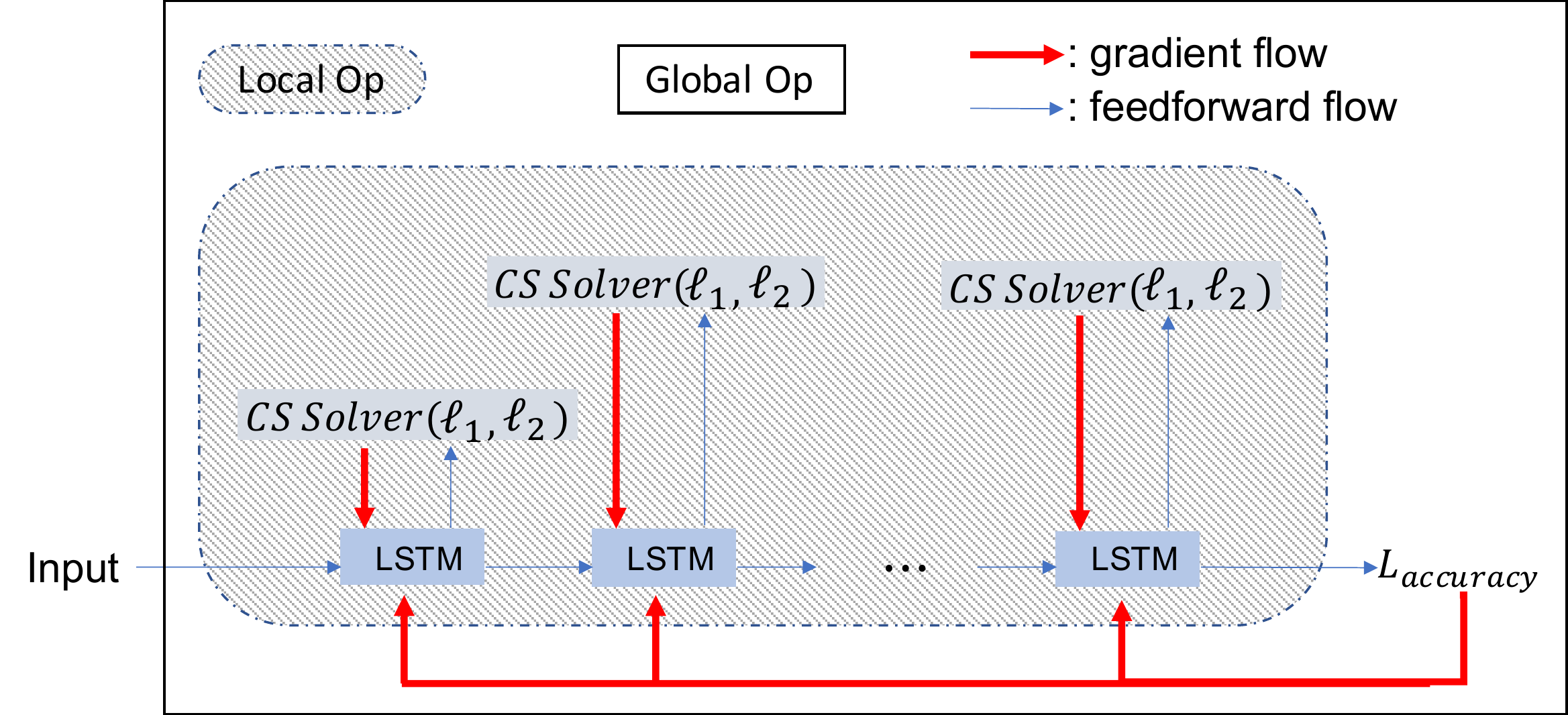}
		\caption{Sparse-aware training scheme for CSP}
		\label{fig:hybrid}
		\vspace{-6mm}
	\end{figure}

	The proposed CSP sparsification optimizer is combined with the conventional backpropagation algorithm in a sparse-aware training scheme (see Fig. \ref{fig:hybrid}). Since CSP is conducted kernel-wise for each layer, the local optimization is triggered during the feedforward stage as the samples passes through the first encoder layer to the last decoder layer sequentially. When the model makes the prediction, the global loss is calculated to update parameters in all preceding layers through backpropagation. The hybrid training scheme is not subjected to gradient vanishing thanks to the local $\ell_1$ regularization, and compatible with global optimization.

\begin{algorithm}[t]
	\caption{Proposed CSP for an LSTM cell}
	\label{algo:1}
	\begin{algorithmic}[1]
		% \SetAlgoLined
		\STATE \textbf{Inputs:} input data at time step $t$, $\bx^{(t)}$\\
		\nonumber  \hspace{0.44in} the hidden state from the previous step, $\bh^{(t-1)}$\\
		\nonumber  \hspace{0.44in} the cell state  from the previous step, $ \bC^{(t-1)}$\\
		%				\nonumber  \hspace{0.4in} the previous hidden and cell states, $[\bh^{(t-1)}, \bh^{(t-1)}]$\\
		%		\nonumber  \hspace{0.4in} the cell state  from the previous step, $ \bC^{(t-1)}$\\
		\STATE \textbf{Outputs:} updated hidden state at time step $t$,  $\bh^{(t)}$\\
		\nonumber  \hspace{0.44in}  updated cell state at time step $t$,  $ \bC^{(t)}$\\
		\IF{sparsity level $<$ the target} 
		\STATE \textbf{Midway inference:} $[\bz_x, \bz_h] = \calF(\mathbb{W}, \bx^{(t)}, \bh^{(t)})$\\
		\STATE \textbf{Sensing:} $\mathbb{W}'\leftarrow\argmin\limits_{\mathbb{W}}\mathcal{L}_{CS}(\mathbb{W}, \bx^{(t)}, \bh^{(t)}, \bz_x, \bz_h)$ 
		%\STATE \textbf{Update pruning threshold} $\rho$ in Eq.\ref{eq:sp}
		\STATE \textbf{Pruning:} $\mathbb{W}_{(i,j)}^p\leftarrow \left \{
		\begin{aligned}
		&0, && \text{if}\ \left| \mathbb{W}'_{(i,j)} \right|<\rho; \\
		&\mathbb{W}'_{(i,j)}, && \text{otherwise}.
		\end{aligned} \right.$
		\STATE \textbf{Update coefficient and threshold}  $\lambda$ and $\rho$
		%\STATE \textbf{Update coefficient}  $\lambda$
		\ENDIF
		\STATE \textbf{Update cell outputs:} $[\bz_x, \bz_h] = \calF(\mathbb{W}^p, \bx^{(t)}, \bh^{(t)})$\\
		\nonumber  \hspace{1.in} $[\bC^{(t)}, \bh^{(t)}] = \calG(\bz_x, \bz_h, \bC^{(t-1)})$ \\
	\end{algorithmic}
\end{algorithm}

	\begin{table*}[t]
		\captionsetup{width=.9\linewidth}
		\caption{Model performance under various sparsity levels for far-field (left) and LibriSpeech (right) datasets.}
		\label{tab:librispeech}
		\begin{subtable}{0.4\textwidth}
			\setlength\tabcolsep{4.0pt}
			\footnotesize
			\centering
			\renewcommand{\arraystretch}{1.1}
			\begin{tabular}{c|c|c|c}
				\hline
				&   & M-I, 38.7M &M-II, 60.0M \\ \hline
				Sparsity (\%)         & Methods     &  \multicolumn{2}{c}{Rel. Dgrd (\%) }   \\ \hline
				0 & -- & --  & -- \\
				\hline
				\multirow{3}{*}{25} & A &   1.83   &        0.80              \\ \cline{2-4} 
				& B &  1.12     &       0.63                  \\ \cline{2-4} 
				& Proposed & \bftab 0.96   &       \bftab0.55                \\ \hline
				\multirow{3}{*}{50} & A &  23.40 &        20.48        \\ \cline{2-4} 
				& B &   8.77   &          7.14         \\ \cline{2-4} 
				& Proposed &   \bftab 6.32    &  \bftab 5.05                     \\ \hline
			\end{tabular}                                                                           
		\end{subtable}
		\begin{subtable}{0.6\textwidth}
			\setlength\tabcolsep{4.pt}
			\footnotesize
			\centering
			\renewcommand{\arraystretch}{1.1}
			\begin{tabular}{c|c|c|c|c|c|c|c}
				\hline			
				&   & \multicolumn{3}{c|}{M-III, 34.0M} & \multicolumn{3}{c}{M-IV, 37.1M} \\ \hline
				Sparsity (\%)             & Methods  & WER  & Abs. Dgrd  & Rel. Dgrd  & WER  & Abs. Dgrd  & Rel. Dgrd\\ \hline
				0                     & --  &    7.27 &     --        &        --     &  9.58 & -- & --    \\ \hline
				\multirow{3}{*}{50} & A &  17.45    &      10.18       &      140.03\%  &  37.24    &      27.66      &      288.73\%     \\ \cline{2-8} 
				& B &   7.34    &  0.07           &      0.96\%   &   10.35   &  0.77           &      8.04\%         \\ \cline{2-8} 
				& Proposed &    \bftab 7.26    &     \bftab -0.01        &      \bftab -0.14\%    &    \bftab 10.06  &     \bftab0.48      &      \bftab 5.01\%       \\ \hline
				\multirow{3}{*}{75} & A &  99.76    &      92.49        &      1272.21\%  &  95.06    &     85.48       &      892.28\%  \\ \cline{2-8} 
				& B &   8.13   &  0.86           &      11.83\%   &     10.43   &  0.85           &     8.87\%       \\ \cline{2-8} 
				& Proposed &    \bftab 8.10    &     \bftab 0.83        &      \bftab 11.42\%     &    \bftab 10.14   &    \bftab  0.56      &      \bftab 5.84\%     \\ \hline
			\end{tabular}
		\end{subtable}
		\vspace{-3mm}
	\end{table*}

	\section{Experiments}
	\label{sec:exp}
	\subsection{Experimental Setup}
	We consider the ASR task with an RNN-T architecture. Comparing to the listen-attend-spell (LAS) model \cite{chan2016listen}, RNN-T features online streaming capability while not requiring a separate lexicon/pronunciation system as in the connectionist temporal classification model (CTC) \cite{graves2006connectionist}.  To rigorously evaluate CSP along with existing sparsification methods, we experiment with 4 different RNN-T based topologies, all with 5 encoding layers and 2 decoding layers. 
	Model M-I, M-III and M-IV all have 768 units per layer (UpL) and 4000 word-pieces (WPs) while M-II uses 1024 UpL with 2500 WPs. Among 4 models, only M-III includes a joint network (J-N), which combines the outputs of RNN-T encoder and decoder to achieve better performance. 
	M-I and M-II are trained on a far-field dataset with 25k hours of English audio, while M-III and M-IV are trained on the LibriSpeech dataset with 960 hours \cite{panayotov2015librispeech}.
	Note that these models are reasonably small comparing with counterparts in the literature, to highlight the effect of sparse pruning on model performance. Please refer to Table \ref{tab:librispeech} for the total number of parameters of each model.

	We compare our proposed CSP with two other pruning methods: na\"{\i}ve pruning and gradual pruning. In the na\"{\i}ve pruning  approach, termed as method-A, the smallest weights are pruned post-training to reach the target sparsity level. 
	Note that for a significantly over-parameterized model, achieving a certain sparsity level with little degradation may not be challenging even with method-A, since most of the weights are not effectively involved in the optimization. 
	Method-A, although not being  spare-aware during training, thus helps us probe  the level of robustness for an RNN-T topology under various sparsity levels. 
	The gradual pruning approach, denoted as method-B, is derived from \cite{zhu2017prune}. 
	%Compared to method-A, gradual pruning provides a mechanism for the model  to recover from the pruning-induced loss. 
	As illustrated in Sec. \ref{sec:approach}, the pruning threshold is calibrated kernel-wise for a fair comparison.
 
	The learning rate in all experiments is specified via a warm-hold-decay scheduler: the initial learning rate, $1e$-$7$, is raised up to $5e$-$4$ at 3K-th step, and is being held till 75K-th step, which then follows a decay stage with which it is reduced to $1e$-$5$ at 200K-th step. The pruning starts at 100K-th step and gradually increases the pruning threshold to reach the target sparsity level at 150K-th step. The intuition, similar to \cite{zhu2017prune}, is to neither apply pruning too early such that the weights are reasonably distributed, nor too late to allow recovering from the sparsification-induced degradation. All models are trained with $10\%$ dropout. The sensing coefficient $\lambda$ initialized as $0.1$. \zhenk{Since $\lambda$ is adjustable according to Eq. \ref{eq:lambda} during training, the results are not predominantly contingent on its initial value. }
	
	\subsection{Experimental Results}
	Consider the performance of M-I and M-II, which do not have a joint network and are trained on the far-field dataset. It is observed in Table \ref{tab:librispeech} that the models are relatively robust at the sparsity level of $25\%$. At $50\%$, the degradation becomes noticeable for all 3 methods. The hard pruning approach does not yield a desirable performance, while our proposed CSP method gives the lowest WER relative degradation in these experimental settings. As expected, the results also indicate a higher relative degradation when the model size decreases. 
	
	In Table \ref{tab:librispeech}, we also report absolute WERs from models trained on the LibriSpeech train dataset and decoded on the LibriSpeech test-clean dataset. Not surprisingly, the models trained with methodA1 also suffer significant degradation at the sparsity level $50\%$. Again, it is observed that the proposed CSP method consistently outperforms all other approaches. Comparing to M-IV,  M-III indicates a higher robustness to pruning likely due to the J-N. At $75\%$ sparsity, all methods experience substantial performance degradation, especially for models with J-N. One reason is that the additional layer actually exacerbates the error, thus leading to higher relative degradation. However, it is worth noting that RNN-T models with J-N still outperforms the counterparts without it by a large margin.
		
	\begin{figure}[t]
		\centering
		\includegraphics[width=0.99\columnwidth]{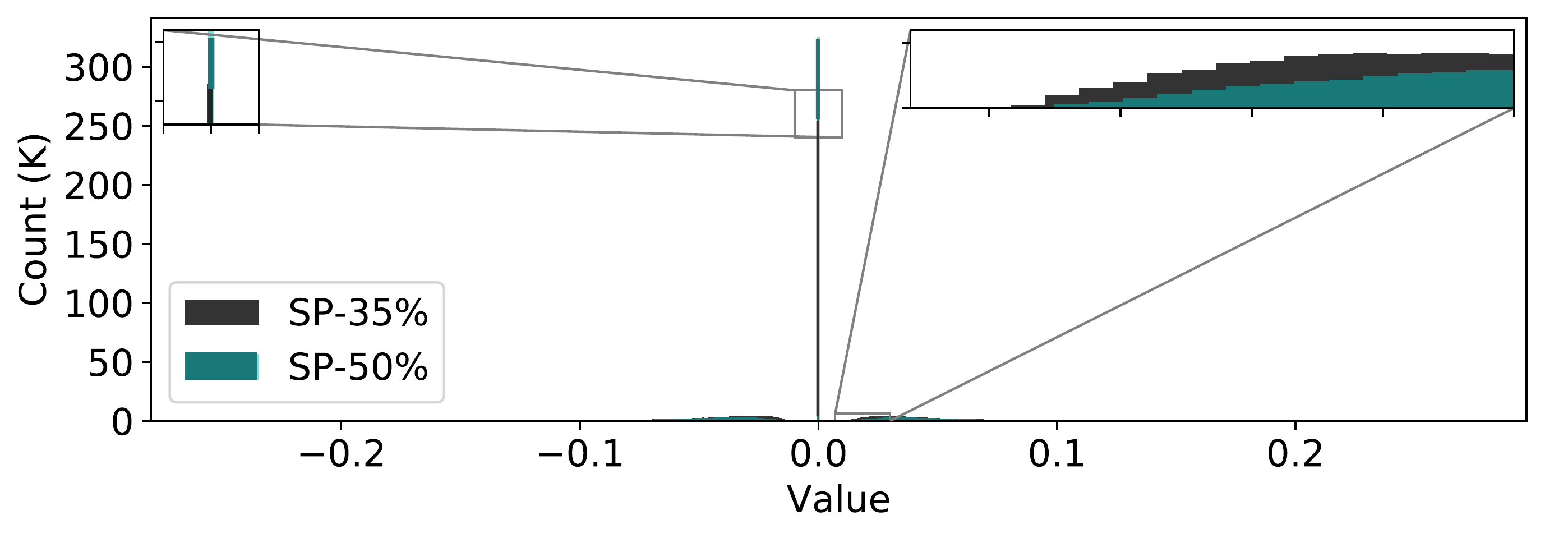}
		%\caption{Redistribution of the weights (a) by CSP from $35\%$ to $50\%$.}
		\caption{Model weight histogram when the sparsity (SP) is up from $35\%$ to $50\%$: the threshold is barely increased with newly pruned weights selected via the CSP method as shown in zoomed-in insets.}
		\label{fig:poc}
		\vspace{-2mm}
	\end{figure}

	\begin{figure}[t]
		\centering
		\includegraphics[width=0.99\columnwidth]{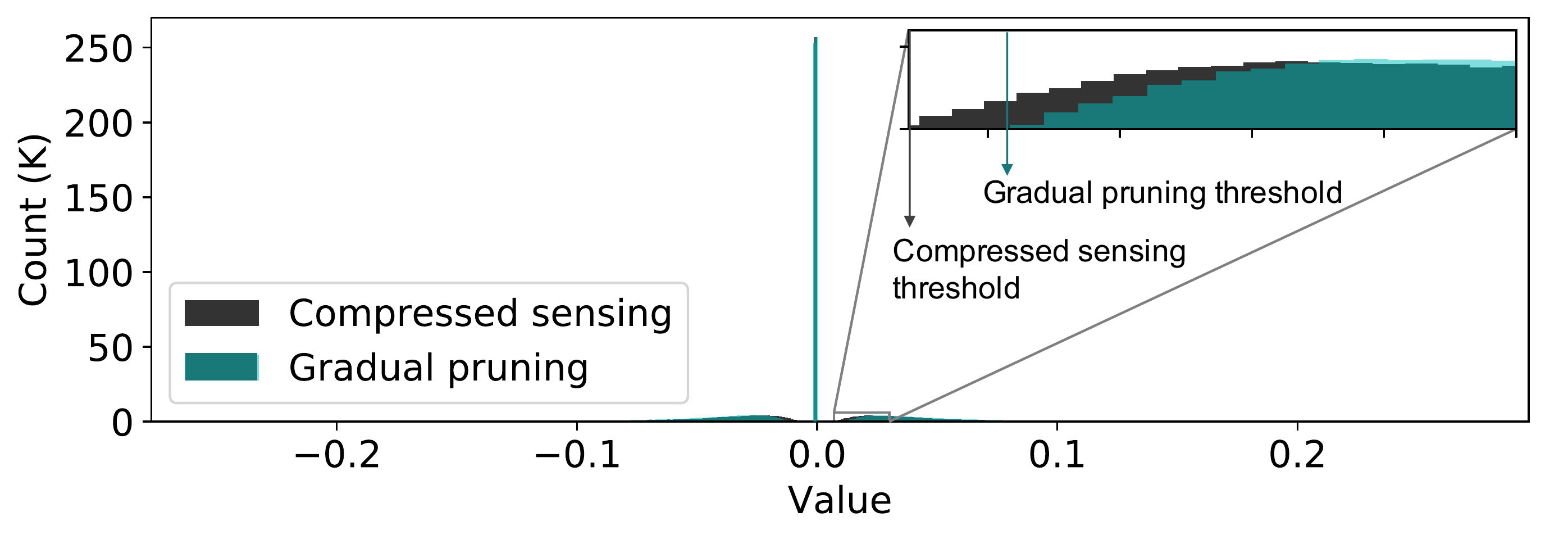}
		\caption{Pruning threshold comparison at $50\%$ sparsity: CSP can redistribute weights to approach the same sparsity level with a smaller pruning threshold than the gradual pruning method.}
		\label{fig:poc2}
		\vspace{-4mm}
	\end{figure}

	%To understand the effect of CSP, we investigate how the model weights are being redistributed when CSP is applied. Fig. \ref{fig:poc} illustrates such process during the M-III model training. To achieve a higher sparsity level of $50\%$ from $35\%$, the pruning threshold does not move by a large margin. Instead, a set of the weights are driven towards 0, and consequently being pruned. In particular, from the top-right zoomed-in figure, we can see that most of the additional weights pruned at $50\%$ are not the ones closest to the threshold at $35\%$ sparsity level. In contrast, gradual pruning approach will significantly increase the pruning threshold to accommodate the higher sparsity level (see Fig. \ref{fig:poc2}). Rather than just hard setting a threshold to achieve sparsity, CSP instead determines ``which (weights) to prune'' via a joint optimization on sparsity and reconstruction regularizers (see Eq.\ref{eq:cs}), thus leading to much smaller WER degradation.
	
	To understand the effect of CSP, we investigate how the model weights are being redistributed when CSP is applied. Fig. \ref{fig:poc} shows the weight distribution when CSP is applied to M-III. It is observed that to achieve a higher sparsity ratio from  $35\%$ to  $50\%$ for CSP, the pruning threshold does not move by a large margin. Instead, a set of the weights are driven towards 0, and consequently being pruned. In particular, from the top-right zoomed-in figure, we can see that most of the additional weights pruned at  $50\%$ in CSP are not those closest to the threshold at  $35\%$ sparsity level. In contrast, the gradual pruning approach will significantly increase the pruning threshold to accommodate the higher sparsity level, and then simply prune the weights with the smallest values (Fig. \ref{fig:poc2}).
	%Please refer to Fig. \ref{fig:poc2}, in which the pruning threshold of gradual pruning is much larger than CSP. 
	Rather than just a hard reset of the threshold, CSP instead determines “which (weights) to prune” via a joint optimization on sparsity and reconstruction regularizers (see Eq.\ref{eq:cs}), thus leading to much smaller WER degradation.
		
%	\begin{figure}[t]
%		\centering
%		\includegraphics[scale=0.25]{hist_2.pdf}
%		\caption{Kernel weight distributions with a sparsity level of ~$50\%$ }
%		\label{fig:hist}
%		\vspace{-4mm}
%	\end{figure}
		
%	Additionally, we find that kernel-wise pruning gives superior performance due to a finer level of pruning granularity. Consider the pruning thresholds of each layer kernel in the RNN-T encoder in Fig.\ref{fig:hist}. With the same sparsity level, the pruning threshold differs among layers. An alternative is to enforce a layer-wise pruning threshold. In that case,  all kernels for one layer share the same pruning threshold which leads to uneven sparsity levels for each kernel. 
%	Thanks to the proposed sensing modulation mechanism, the kernel-wise CSP does not introduce redundant sparsity, which mitigates the model degradation.
%	We omit the visualization for weights of recurrent kernels and decoder, as they follow a similar pattern. 

	\section{Conclusions}
	\label{sec:conclusion}
	We propose a novel pruning approach for machine learning model compression based on compressed sensing, termed as CSP. Compared to existing sparsification methods which focus only on ``when to prune'', CSP further addresses the question ``which (weights) to prune'' by considering both sparsity inducing and compression-error reduction mechanisms. We validate the effectiveness of CSP via the speech recognition task with RNN-T model. CSP achieves superior results compared to other sparsification approaches. The proposed method can be straightforwardly incorporated into other ML models and/or compression methods to further reduce model complexity.

	\newpage
	
	% References should be produced using the bibtex program from suitable
	% BiBTeX files (here: strings, refs, manuals). The IEEEbib.bst bibliography
	% style file from IEEE produces unsorted bibliography list.
	% -------------------------------------------------------------------------
	\bibliographystyle{IEEEbib}
	\balance
	\bibliography{strings,refs}

\end{document}